\documentclass[12pt, twoside]{article}

\usepackage[a4paper, total={5.38in, 8.6in}]{geometry}

\usepackage{titlesec}

\makeatletter
\titleformat{\section}
       {\large\bfseries}
       {\@seccntformat{section}}
       {0.5em}
       {}
       []

\titleformat{\subsection}[runin]
       {\normalfont\bfseries}
       {\thesubsection}
       {0.5em}
       {}
       [.]

\titleformat{\subsubsection}[runin]
       {\normalfont\itshape}
       {\thesubsubsection}
       {0.5em}
       {}
       [.]

\titleformat{\paragraph}[runin]
       {\normalfont\itshape}
       {}
       {0.5em}
       {}
       [.]

\usepackage{amsfonts}       
\usepackage{amssymb,amsmath}
\usepackage{nicefrac}       
\usepackage{microtype}      
\usepackage{graphicx}
\usepackage{textcomp}
\usepackage{xcolor}
\usepackage{comment}
\usepackage{caption}
\usepackage{makecell} 
\usepackage{multirow}
\usepackage{flushend}
\usepackage{mathtools}
\usepackage[sort, authoryear]{natbib}
\usepackage{enumitem}
\setlist{nolistsep}
\usepackage[title]{appendix}

\usepackage{setspace}

\usepackage{algorithm}
\usepackage{algpseudocode}
 \usepackage{setspace}

\usepackage{fancyhdr}

\makeatletter
\renewcommand{\maketitle}{\bgroup\setlength{\parindent}{0pt}
\begin{flushleft}
  \textbf{\@title}

  \@author
\end{flushleft}\egroup
}
\makeatother

\DeclarePairedDelimiter\norm{\lVert}{\rVert}
\DeclareMathOperator*{\argmax}{arg\,max}

\numberwithin{equation}{section}

\begin{document}

    \fancyhead{}
    \fancyfoot{}
    \fancyhead[LE, RO]{\small \thepage}
    \fancyhead[RE]{\small D. Kleyko et al.}
    \fancyhead[LO]{\small Efficient Decoding of Compositional Structure in Holistic Representations}
    \pagestyle{fancy}

\title{\Large \textbf{Efficient Decoding of Compositional Structure in Holistic Representations}
}

\author{
\phantom{text} \\
\textbf{{Denis~Kleyko}}\\
\textit{\mbox{denis.kleyko@ri.se}} \\
\textit{Redwood Center for Theoretical Neuroscience, University of California at Berkeley, Berkeley, CA 94720, U.S.A., and Intelligent Systems Laboratory, Research Institutes of Sweden, 16440 Kista, Sweden}\\
\phantom{text} \\
\textbf{{Connor~Bybee}}\\
\textit{\mbox{bybee@berkeley.edu}} \\
\textbf{{Ping-Chen~Huang}}\\
\textit{\mbox{pingchen.huang@berkeley.edu}} \\
\textbf{{Christopher J. Kymn}}\\
\textit{\mbox{cjkymn@berkeley.edu}} \\
\textbf{{Bruno A. Olshausen}}\\
\textit{\mbox{baolshausen@berkeley.edu}} \\
\textit{Redwood Center for Theoretical Neuroscience, University of California at Berkeley, Berkeley, CA 94720, U.S.A.}\\
\phantom{text} \\
\textbf{{E.~Paxon~Frady}}\\
\textit{\mbox{e.paxon.frady@intel.com}} \\
\textit{Neuromorphic Computing Laboratory, Intel Labs, Santa Clara, CA 95054, U.S.A.}\\
\phantom{text} \\
\textbf{{Friedrich~T.~Sommer}}\\
\textit{\mbox{fsommer@berkeley.edu}} \\
\textit{Redwood Center for Theoretical Neuroscience, University of California at Berkeley, Berkeley, CA 94720, U.S.A., and Neuromorphic Computing Laboratory, Intel Labs, Santa Clara, CA 95054, U.S.A.}\\
}

\date{}

\maketitle

\thispagestyle{empty}

\noindent
\textbf{We investigate the task of retrieving information from compositional distributed representations formed by Hyperdimensional Computing/Vector Symbolic Architectures and present novel techniques which achieve new information rate bounds.
First, we provide an overview of the decoding techniques that can be used to approach the retrieval task.
The techniques are categorized into four groups.
We then evaluate the considered techniques in several settings  that involve, e.g., inclusion of external noise and storage elements with reduced precision. 
In particular, we find that the decoding techniques from the sparse coding and compressed sensing literature (rarely used for Hyperdimensional Computing/Vector Symbolic Architectures) are also well-suited for decoding information from the compositional distributed representations.
Combining these decoding techniques with interference cancellation ideas from communications improves previously reported bounds~\citep{HerscheHDMClassifier2021} of the information rate of the distributed representations from 1.20 to 1.40 bits per dimension for smaller codebooks and from 0.60 to 1.26 bits per dimension for larger codebooks.  }

\section{Introduction}

Hyperdimensional Computing~\citep{KanervaHyperdimensional2009} a.k.a. Vector Symbolic Architectures~\citep{GaylerJackendoff2003} (HD/VSA) allows the formation of rich, compositional, distributed representations that can construct a plethora of data structures~\citep{KleykoComputingParadigm2021, demidovskij2021encoding}.
Although each individual field of a data structure is encoded in a fully distributed manner, it can be decoded (and manipulated) individually. 
This decoding property provides the remarkable transparency of HD/VSA, in stark contrast to the opacity of traditional neural networks~\citep{Shwartz2017opening}.
For example, decoding of distributed representations enables the tracing (or explanation) of individual results. 
It even led to the proposal of HD/VSA as a programming framework for distributed computing hardware~\citep{KleykoComputingParadigm2021}. 
However, there are capacity limits on the size of data structures that can be decoded from fixed-sized distributed representations, and these limits depend on the decoding techniques that are used. 
Here, we will characterize different techniques for decoding information from distributed representations formed by HD/VSA and provide empirical results on the information rate, including results for novel decoding techniques.
The reported results are interesting from a theoretical perspective, as they exceed the capacity limits previously thought to hold for distributed representations~\citep{FradyCapacity2018,HerscheHDMClassifier2021}.
From a practical perspective, many applications of HD/VSA hinge on efficiently decoding information stored in distributed representations, including, e.g., communications~\citep{ JakimovskiCollective2012, KleykoMACOM2012, KimHDM2018, HsuNonOrthogonalModulation2020, WirelessGuirado2022} and distributed orchestration~\citep{SimpkinHDWorkflow2019}.

The problem of decoding information from distributed representations has similarities to information retrieval problems in other areas, such as in communications, reservoir computing, sparse coding, and compressed sensing. 
Here, we will describe how techniques developed in these areas can be applied to HD/VSA.
The major contributions of the study are: 
\begin{itemize}
    \item A taxonomy of decoding techniques suitable for retrieval from representations formed by HD/VSA;
    
    \item A comparison of ten decoding techniques on a retrieval task; 
    
    \item A qualitative description of the tradeoff between the information capacity of distributed representations and the amount of computation the decoding requires;
    
    \item Improvements on the known bounds on information capacity for distributed representations of data structures (in bits per dimension)~\citep{FradyCapacity2018,HerscheHDMClassifier2021}. 
    
\end{itemize}

The paper is structured as follows. 
Section~\ref{sect:decoding} introduces the approaches suitable for decoding from distributed representations. 
The empirical evaluation of the introduced decoding techniques is reported in Section~\ref{sect:empirical}. 
The findings are discussed in Section~\ref{sect:discussion}.

Readers interested in further background for HD/VSA are encouraged to read Appendix~\ref{sect:supp:vsa}. 
To evaluate the considered decoding techniques, we consider the case of an $n$-dimensional vector, $\mathbf{y}$, that represents a sequence of symbols $\mathbf{s}$ of length $v$. The symbols are drawn randomly from an alphabet of size $D$.
Symbols are represented by $n$-dimensional random bipolar vectors that are stored in the codebook $\mathbf{\Phi}$.  
The permutation and superposition operations are used to form $\mathbf{y}$ from representations of sequence symbols $\mathbf{\Phi}_{\mathbf{s}_i}$.
We then use the decoding techniques to construct $\hat{\mathbf{s}}$, a reconstruction of $\mathbf{s}$ using $\mathbf{y}$ and the codebook $\mathbf{\Phi}$. 
For further details on the encoding scheme, please see Appendix~\ref{sec:problem}. 
We evaluate the quality of $\hat{\mathbf{s}}$ based on accuracy and information rate; further details are provided in Appendix~\ref{sec:back:perf}.

\section{Decoding techniques}
\label{sect:decoding}

\begin{figure}[t]
    \centering
    \includegraphics[width=1.0\columnwidth]{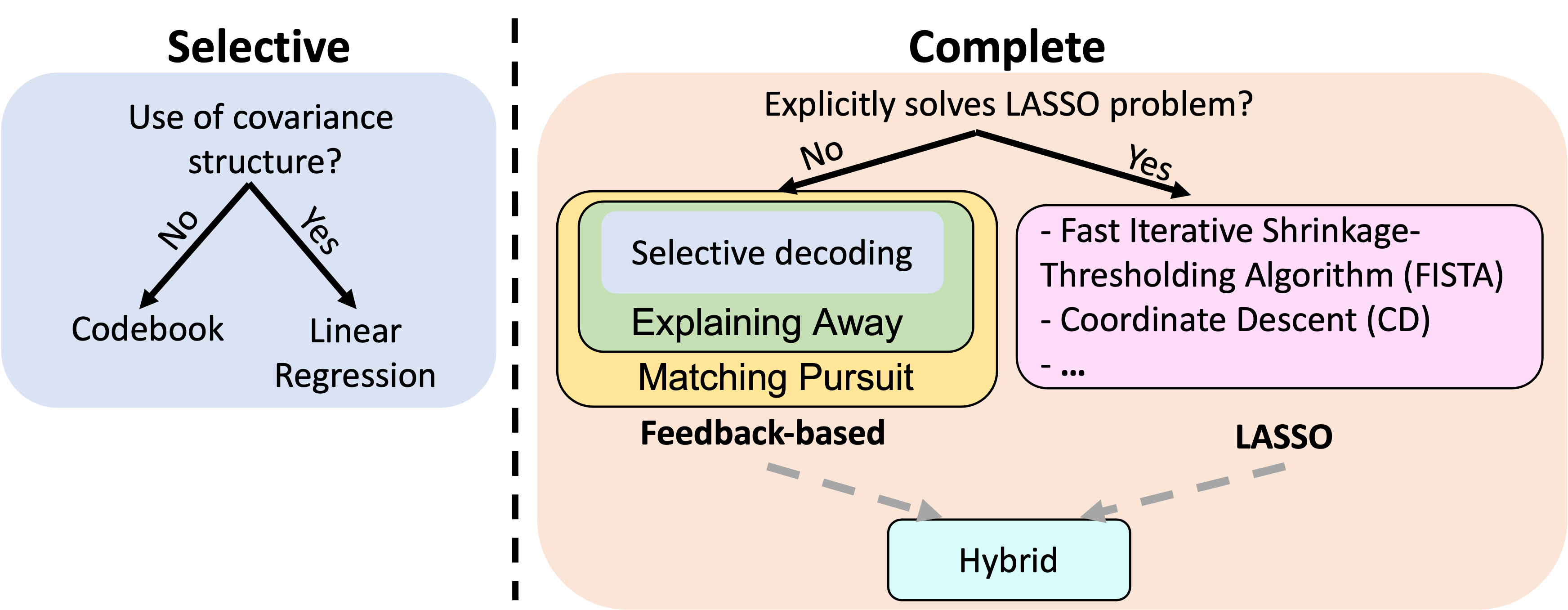}
    \caption{
    The taxonomy of decoding techniques surveyed and investigated. 
    Techniques for selective decoding require two steps: first, the transformation of the query vector to select a particular field and second, a simple feed-forward matrix-vector multiplication, followed by $\argmax$ (Section~\ref{sect:decoding:readout}). In contrast, techniques for complete decoding (Section~\ref{sect:decoding:complete}) do not retrieve a single field, but rely on iterative procedures that are computationally more intensive.
    Ellipses indicate that there are additional techniques within a group that are not considered here.
    }
    \label{fig:taxonomy}
\end{figure}

In this section, we survey decoding techniques for retrieving sequence symbols (denoted as $\hat{\mathbf{s}}$) from their compositional distributed representation  $\mathbf{y}$ (see Appendix~\ref{sec:problem} for a detailed problem formulation). 
The decoding techniques can be taxonomized into two types, {\it selective} (Section~\ref{sect:decoding:readout}) and {\it complete} (Section~\ref{sect:decoding:complete}).
In selective decoding, a query input selects a particular field in the data structure, which is then decoded individually. 
Conversely, in complete decoding, all fields of the data structure are decoded simultaneously.
In the following subsections, we introduce concrete decoding techniques pertaining to the two different types. 
The taxonomy of these decoding techniques and their relationships are summarized in Figure~\ref{fig:taxonomy}.

\subsection{Techniques for selective decoding}
\label{sect:decoding:readout}

In techniques for selective decoding, a query input selects a particular field of the data structure represented by a distributed representation (vector $\mathbf{y}$). The content of the selected data field $\hat{\mathbf{s}}_i$ is then decoded.
In techniques for selective decoding, information about the field $i$ of the query in the data structure   
is translated into a readout matrix (denoted as $\mathbf{W}^{\mathrm{out}}(i)  \in [D \times n]$), which is then used for decoding the data field. 
We adopt here the term readout matrix from the reservoir computing literature~\citep{LukoseviciusRC2009}. In reservoir computing, many readout matrices can be specified for a single distributed representation depending on the task. 
In the framework of HD/VSA, the different readout matrices correspond to queries of different fields in the data structure represented by the compositional distributed representation $\mathbf{y}$.

Once the readout matrices are known, a prediction for symbol in position $i$ is computed as: 
\noindent
\begin{equation*}
\hat{\mathbf{s}}_i = \argmax (\mathbf{W}^{\mathrm{out}}(i) \mathbf{y}  ),
\end{equation*}
\noindent
where $\argmax(\cdot)$ returns the symbol with the highest dot product.
Thus, selective techniques make their predictions by choosing a symbol corresponding to a row in $\mathbf{W}^{\mathrm{out}}(i)$ that has the highest dot product with $\mathbf{y}$.  
Below, we consider two ways of forming $\mathbf{W}^{\mathrm{out}}(i)$.

\subsubsection{Codebook decoding}

``Codebook decoding'' is the simplest technique for selectively decoding $\hat{\mathbf{s}}$ from $\mathbf{y}$. 
It corresponds to a matrix-vector multiplication between the query high-dimensional vector (a.k.a. hypervector) such as $\mathbf{y}$ and algorithm's codebook, a matrix containing all hypervectors representing symbols. 
Here, the codebook $\mathbf{\Phi}  \in \{-1,1\}^{n \times D}$ stores atomic $n$-dimensional random i.i.d. bipolar hypervectors assigned to $D$ unique atomic symbols of the alphabet.
Because this technique of decoding is omnipresent in HD/VSA; we call it ``Codebook decoding''. 
Further, we use a permutation with long cycle length (denoted as $\rho^i(\cdot)$) to bind a symbol with its position in the sequence (see details in Appendix~\ref{sec:problem}). 
For example, to represent the $i$th position, the corresponding entry of the codebook $\mathbf{\Phi}_{\mathbf{s}_i}$ can be permuted $v-i$ times. 
Then, the readout matrix for the symbol at position $i$ in the sequence is constructed as:
\noindent
\begin{equation}
\mathbf{W}^{\mathrm{out}}(i)= \rho^{v-i} (\mathbf{\Phi}^\top).
\label{eq:readout:VSA}
\end{equation}
\noindent
Note that an equivalent way to decode is by applying an inverse permutation operation (denoted as $\rho^{-1}$) $v-i$ times to $\mathbf{y}$ without permuting $\mathbf{\Phi}$: 
\noindent
\begin{equation}
\hat{\mathbf{s}}_i =  \argmax \left ( \rho^{v-i} (\mathbf{\Phi}^\top) \mathbf{y} \right  ) 
=\argmax \left (\mathbf{\Phi}^\top \rho^{-(v-i)} ( \mathbf{y} )  \right).
\end{equation}
\noindent
This formulation is more practical, as only a single vector (rather than a matrix) is being permuted. 
Finally, it is worth noting that the performance of Codebook decoding could be predicted analytically using the methodologies from~\cite{FradyCapacity2018}.
This point was confirmed in our experiments below (see Figures~~\ref{fig:perf:noiseless}-\ref{fig:perf:dimen}), where the gray dashed lines depicting the analytical predictions closely match our experiments.

\subsubsection{Linear regression decoding}
\label{sect:decoding:readout:cov}

Note that the hypervectors in $\mathbf{\Phi}$ are not perfectly orthogonal to each other. 
Therefore, Codebook decoding benefits from adjustments based on the expected covariance matrix of hypervectors interacting in $\mathbf{y}$ (denoted as $\tilde{\mathbf{C}}$):
\noindent
\begin{equation}
\mathbf{W}^{\mathrm{out}}(i)= \left( \tilde{\mathbf{C}}^{-1} \rho^{v-i} (\mathbf{\Phi})  \right) ^\top.
\label{eq:readout:cov}
\end{equation}
\noindent
The readout matrix in (\ref{eq:readout:cov}) does not differ much from the one in (\ref{eq:readout:VSA}) used for Codebook decoding, yet it substantially improves the information rate for small $D$ (Section~\ref{sect:empirical:noiseless}). 
Nevertheless, this technique does not seem to be widely known within the HD/VSA literature, hence, this study aims at placing it on the map for those in the area. Below, we will abbreviate this technique as ``LR decoding''.   

For the considered transformation (Eq.~(\ref{eq:perm_seq}) in~\ref{sec:problem}), $\tilde{\mathbf{C}}$ can be calculated analytically~\citep{FradyCapacity2018} as:  
\noindent
\begin{equation}
\begin{split}
\tilde{\mathbf{C}} = & \sum_{i=1}^{v} \rho^{i} (\mathbf{\Phi}) \left( \frac{1}{D}\mathbf{I} \right) \rho^{-i} (\mathbf{\Phi})^\top  + \\
&+\sum_{i=1}^{v} \sum_{j=1; j\neq i}^{v} \rho^{i} ( \mathbf{\Phi}) \left( \frac{1}{D^2}\mathbf{J} \right) \rho^{-j} (\mathbf{\Phi})^\top,
\end{split}
\label{eq:cov:fine}
\end{equation}
\noindent
where $\mathbf{I} \in [D \times D]$ is the identity matrix while $\mathbf{J} \in [D \times D]$ is the unit matrix.
Note that compared to Codebook decoding, LR decoding incurs additional computational costs for obtaining $\tilde{\mathbf{C}}^{-1}$ and computing $\mathbf{W}^{\mathrm{out}}(i)$ that scale with $n$, $D$, and, $v$.
In general, however, the covariance-based readout matrix cannot be computed analytically. 
Therefore, the standard practice in the randomized neural networks literature~\citep{LukoseviciusRC2009,RCNNSsurvey, KleykoDensityEncoding2020} is to collect some training data and use minimum mean square error to estimate the optimal values of the  readout matrix by solving a linear regression (LR) problem.

\subsection{Techniques for complete decoding}
\label{sect:decoding:complete}

Since hypervectors in $\mathbf{\Phi}$ and their permuted versions are not completely orthogonal, summing them in $\mathbf{y}$ produces crosstalk noise that degrades the result of the selective decoding.
Therefore, there are techniques attempting the complete decoding of the data structure, for example, by first selectively decoding all fields of the data structure, and then using the decoding results to remove crosstalk noise introduced by other fields and repeating the selective decoding.
Below, we overview three kinds of techniques: feedback-based,
Least Absolute Shrinkage and Selection Operator (LASSO), and hybrid (combining elements of both). 

Prior to introducing the techniques for complete decoding, we can make an interesting observation by reconsidering the transformation used to form a sequence's compositional hypervector $\mathbf{y}$ (Eq.~(\ref{eq:perm_seq}) in~\ref{sec:problem}).
Similar to readout matrices for Codebook decoding in (\ref{eq:readout:VSA}), we can make $v$ permuted versions of the codebook $\mathbf{\Phi}$ and concatenate them horizontally into one large codebook (denoted as $\mathbf{A} \in [n \times vD]$) as:
\noindent
\begin{equation}
\mathbf{A}= [ \rho^{v-1}( \mathbf{\Phi}), \rho^{v-2}( \mathbf{\Phi}),   \cdots, \rho^{1}( \mathbf{\Phi}), \rho^{0}( \mathbf{\Phi})].
\label{eq:reg:codebook}
\end{equation}
\noindent
Next, we can form a vector $\mathbf{x} \in [vD \times 1]$ that will contain the concatenation of $D$-dimensional one-hot encodings of symbols in the original sequence $\mathbf{s}$ (denoted as $\mathbf{o}_{\mathbf{s}_i}$):
\noindent
\begin{equation}
\mathbf{x}= [\mathbf{o}_{\mathbf{s}_1}^\top, \mathbf{o}_{\mathbf{s}_2}^\top,   \cdots, \mathbf{o}_{\mathbf{s}_v}^\top]^\top.
\label{eq:reg:onehot}
\end{equation}
\noindent
Note that multiplication of $\mathbf{A}$ by $\mathbf{x}$ is now equivalent to Eq.~(\ref{eq:perm_seq}) in~\ref{sec:problem}:
\noindent
\begin{equation}
\label{eq:reg:sup}
\mathbf{A} \mathbf{x}= \mathbf{y} = \sum_{i=1}^{v}\rho^{v-i}( \mathbf{\Phi}_{\mathbf{s}_i}).
\end{equation}
\noindent
In this formulation, the complete decoding can be seen as an optimization problem of finding $\hat{\mathbf{x}}$ for given $\mathbf{A}$ and $\mathbf{y}$ such that:
\noindent
\begin{equation}
    \hat{\mathbf{x}}=\min_{\mathbf{x}} \norm{\mathbf{A} \mathbf{x} - \mathbf{y}}_2.
    \label{eq:reg:opt}
\end{equation}
\noindent
Below, we consider several approaches to solving (\ref{eq:reg:opt}).

\subsubsection{Feedback-based techniques}
\label{sect:decoding:feedback}

The key idea of feedback-based techniques is to leverage initial predictions $\hat{\mathbf{s}}$ (obtained, e.g., with one of the selective techniques from Section~\ref{sect:decoding:readout}) to remove crosstalk noise in the decoding of one field by subtracting the hypervectors for all (or some) other fields in~$\mathbf{y}$. 
Similar ideas for such a feedback mechanism have been developed in other fields of research and referred to as: ``explaining away'', ``interference cancellation'', or ``peeling decoding''.
Below, we consider two feedback-based decoding techniques: explaining away (EA) feedback and matching pursuit with explaining away.

\paragraph{Explaining away}

To reduce the crosstalk noise in decoding one field $i$ of the data structure, this technique constructs the corresponding hypervector from the decoding predictions of all other data fields and subtracts it from $\mathbf{y}$.
Under the assumption that most of the decoding predictions $\hat{\mathbf{s}}$ are correct, this subtraction significantly reduces crosstalk for decoding the data field $i$. 
Formally, this can be written as:
\noindent
\begin{equation}
\hat{\mathbf{y}}(i) = \mathbf{y} - \sum_{j=1; j\neq i}^{v}\rho^{v-j}( \mathbf{\Phi}_{\hat{\mathbf{s}}_j}).
\label{eq:feedback:ea}
\end{equation}
\noindent
The hypervector $\hat{\mathbf{y}}(i)$ with reduced crosstalk noise is then used with the corresponding readout matrix $\mathbf{W}^{\mathrm{out}}(i)$ to revise the prediction for the $i$th position of $\hat{\mathbf{s}}$:
\noindent
\begin{equation}
\hat{\mathbf{s}}_i = \argmax (\mathbf{W}^{\mathrm{out}}(i) \hat{\mathbf{y}}(i)  ).
\label{eq:prediction:ea}
\end{equation}
\noindent
This process is repeated iteratively until either the predictions in $\hat{\mathbf{s}}$ stop changing or the maximum number of iterations (denoted as $r$) is reached. 
We consider two variants of EA using the two variants of selective decoding described above: ``Codebook EA'' and ``LR EA''.

\paragraph{Matching pursuit with explaining away}
\label{sect:decoding:complete:feed:mp}

One issue with EA is that when many of the decoding predictions in $\hat{\mathbf{s}}$ are wrong, the subtraction adds rather than removes noise. 
One possibility to counteract this problem is by successively subtracting individual decoded fields in the hypervector, starting with the ones for which the confidence of the correct decoding is highest.
As a confidence measure for selective decoding, we choose the cosine similarity between the (residual) hypervector and the best, appropriately permuted, matching codebook entry.
A {\it confidence score} is calculated as the difference between the highest and the second highest cosine similarities. 
Intuitively, we expect that a high confidence score should correlate with the decoding result being correct.

Using a technique for selective decoding for all positions, we choose the decoding result in position $c$ with the highest  
confidence score and remove this prediction $\hat{\mathbf{s}}_c$ from the compositional hypervector $\mathbf{y}$: 
\noindent
\begin{equation}
\tilde{\mathbf{y}}= \mathbf{y} - \rho^{v-c}( \mathbf{\Phi}_{\hat{\mathbf{s}}_c}).
\label{eq:feedback:confidence}
\end{equation}
\noindent
From now on, the prediction for position $c$ is fixed, and we assume that $\tilde{\mathbf{y}}$ stores only $v-1$ symbols\footnote{
Note that it is still important to keep track of the original positions in the sequence due to the use of position-dependent permutations. 
}.
The new hypervector $\tilde{\mathbf{y}}$ can be used with EA to make new predictions for the remaining $v-1$ symbols. 
Then, we choose the most confident prediction among these $v-1$ symbols, fix the prediction, remove it from $\tilde{\mathbf{y}}$, and repeat the EA decoding for the remaining $v-2$ symbols.
In such a manner, the decoding proceeds successively until complete. 
This type of confidence-based EA is similar to matching pursuit (MP), a well-known greedy technique for sparse signal approximation~\cite{MallatMatching1993}. 
In a step of MP, the best matching codebook element is weighted with the dot product between signal and codebook element to explain as much as possible of the signal. 
The next MP step continues on the residual. 
Essentially, (\ref{eq:feedback:confidence}) is also the residual of an MP approximation. 
However, the goal here is to explain away an element of the hypervector representing one field of the data structure. 
As the encoding procedure weights all used codebook elements with a value of one, the weight chosen in the residual is also one.  
Similar to the case of EA, below, we will investigate MP decoding with the two variants of selective decoding: ``Codebook MP'' and ``LR MP''.

\subsubsection{LASSO techniques}
\label{sect:decoding:optimization}
The formulation in Eq.~(\ref{eq:reg:opt}) can be conceptualized as trying to infer a solution simultaneously, i.e., trying to decode the whole data structure at once. 
Note that this problem formulation is a relaxed version of the original task, as it does not take into account the constraint that there is only one nonzero component within each $D$-dimensional segment of $\hat{\mathbf{x}}$. 
We can simply impose this constraint and form $\hat{\mathbf{s}}$ from $\hat{\mathbf{x}}$ by assigning $\hat{\mathbf{s}}_i$ to the position of the highest component of $i$th $D$-dimensional segment of $\hat{\mathbf{x}}$.

Another way to think about the constraint above is that it is as if we have prior knowledge that $\hat{\mathbf{x}}$ has only $v$ nonzero components.
This means that the density of $\hat{\mathbf{x}}$ should be $1/D$ so the expected solution becomes quite sparse even for moderately large values of $D$. 
Therefore, the problem in (\ref{eq:reg:opt}) can be treated as the sparse inference procedure used in the areas of sparse coding~\citep{OlshausenEmergence1996} and compressed sensing~\citep{DonohoCompressed2006}. 
Thus, the natural choice for techniques to solve (\ref{eq:reg:opt}) should come from an arsenal of methods developed within the sparse coding/compressed sensing literature. 
The most common approach to do so is via a well-known LASSO regression~\citep{TibshiraniRegression1996} that adds L1 norm regularization to (\ref{eq:reg:opt}):
\noindent
\begin{equation}
    \hat{\mathbf{x}}=\min_{\mathbf{x}} \norm{\mathbf{A} \mathbf{x} - \mathbf{y}}_2 + \lambda \norm{\mathbf{x}}_{1},
    \label{eq:reg:lasso}
\end{equation}
\noindent
where $\lambda$ is a standard hyperparameter denoting the importance of the L1 regularization term. Coordinate Descent (CD) and gradient descent algorithms are investigated for solving the LASSO. CD is implemented by \cite{scikit-learn} (``CD decoding'') and the Fast Iterative Shrinkage-Thresholding Algorithm (FISTA, ``FISTA decoding'')~\citep{BeckFISTA2009fast} is used for gradient descent.

\subsubsection{Hybrid techniques}
\label{sect:decoding:hybrid}

Hybrid techniques combine primitives from the previous techniques as indicated by the dashed arrows in Figure~\ref{fig:taxonomy}.  
Although there is no fixed recipe for combining techniques together, we show that one particularly powerful technique is in combining CD or FISTA decoding and LR decoding with MP (``CD/LR MP'' \& ``FISTA/LR MP'').
In these techniques, either CD or FISTA decoding is used every time when the current most confident prediction is explained away from $\mathbf{y}$ according to (\ref{eq:feedback:confidence})\footnote{
CD or FISTA decoding is also used to make the initial predictions from the original $\mathbf{y}$.
}
while LR EA decoding is used to improve CD's or FISTA's predictions for the symbols that are not yet fixed.

\section{Empirical evaluation}
\label{sect:empirical}

In the experiments, we focus on three settings for the decoding: 
\begin{itemize}
    \item Decoding in the absence of external noise (Section~\ref{sect:empirical:noiseless});
    \item Decoding in the presence of external noise (Section~\ref{sect:empirical:noisy});
    \item Decoding from storage elements with limited precision
    (Section~\ref{sect:results:clipping}).
\end{itemize}
\noindent
Before going into the results of evaluation let us briefly repeat the notations introduced so far as they will be used intensively below. 
$\mathbf{y}$ is an $n$-dimensional vector that represents a sequence $\mathbf{s}$ of $v$ symbols where the symbols are chosen from an alphabet of size $D$ whose representations are stored in the codebook $\mathbf{\Phi}$.  
A hypervector of an $i$-th symbol in $\mathbf{s}$ is denoted by $\mathbf{\Phi}_{\mathbf{s}_i}$ while the reconstructed sequence is denoted as $\hat{\mathbf{s}}$.

\subsection{Noiseless decoding}
\label{sect:empirical:noiseless}

\begin{figure}[t]
    \centering
    \includegraphics[width=1.0\columnwidth]{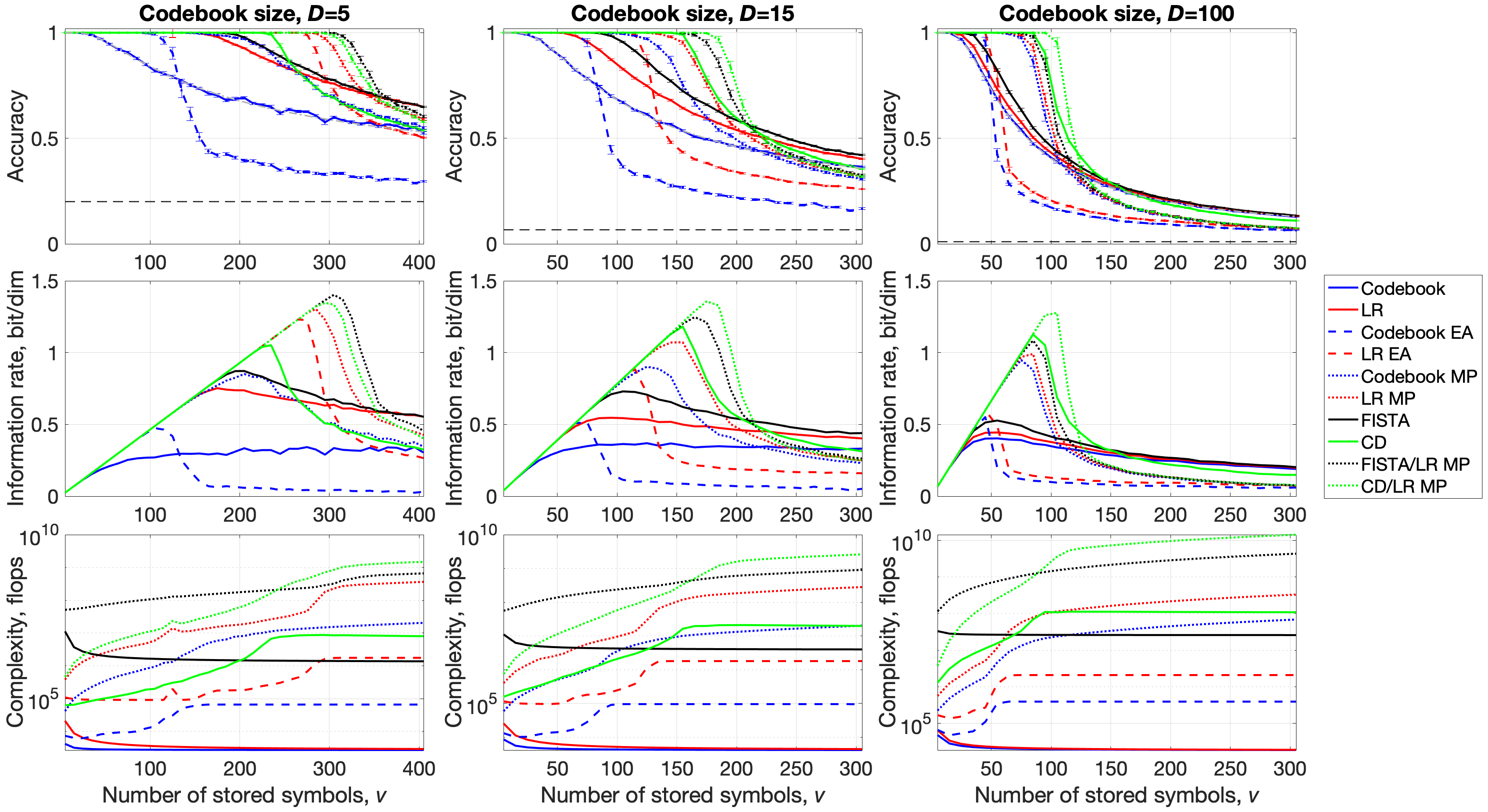}
    \caption{
    The decoding accuracy, information rate, and computational complexity against sequence length $v$ for three different codebook sizes $D$. 
    No external noise was added to $\mathbf{y}$.
    The upper panels depict accuracy, the middle panels correspond to information rate, while the lower panels show complexity in terms of floating point operations. 
    The reported results are averages obtained from twenty randomly initialized codebooks. 
    Ten random sequences were simulated for each codebook per each $v$.
    In the upper panels, bars depict $95\%$ confidence intervals, thin dashed black lines indicate the corresponding random guess at $1/D$; while gray dashed lines correspond to analytical predictions for Codebook decoding.  
    }
    \label{fig:perf:noiseless}
\end{figure}

We begin by comparing the surveyed decoding techniques in Section~\ref{sect:decoding} in a scenario when no external noise is added to $\mathbf{y}$. This follows the setup of ~\citep{HerscheHDMClassifier2021}, which previously reported the highest information rate of HD/VSA in bits per dimension that could be achieved in practice.
In the experiments in~\citep{HerscheHDMClassifier2021}, $n$ was set to $500$ (see Appendix~\ref{sect:dimensionality} that reports the effect of $n$); $D$ was chosen from $\{5, 15,100\}$ and $v$ varied between $0$ and $300$ (we used $400$ for $D=5$). 
The results of the experiments for the techniques from Section~\ref{sect:decoding} are presented in Figure~\ref{fig:perf:noiseless}. 
In~\citep{HerscheHDMClassifier2021}, only the first four out of ten techniques (see legend in Figure~\ref{fig:perf:noiseless}) compared here\footnote{
It also reported a ``soft-feedback'' technique that we do not report here due to its high similarity to EA. 
}
were considered.
The best information rate (see definition in Eq.~(\ref{eq:MI:dim}) Appendix in~\ref{sec:information}) achieved in~\citep{HerscheHDMClassifier2021}, was approximately $1.20$, $0.85$, and $0.60$ bits per dimension for $D$ equal to $5$, $15$, and $100$, respectively. 
The key takeaway from Figure~\ref{fig:perf:noiseless} is the improvement over previously achieved information rates.
The new highest results for information rate are $1.40$ ($17$\% improvement), $1.34$ ($58$\% improvement), and $1.26$ ($110$\% improvement) bits per dimension for $D$ equal to $5$, $15$, and $100$, respectively. 

Note that for all values of $D$, the highest information rate was obtained with the hybrid techniques.
This matches the fact that the LASSO techniques alone demonstrated high-fidelity (i.e., close to perfect) regimes of decoding accuracy (see definition in Eq.~(\ref{eq:accuracy}) in Appendix~\ref{sec:back:perf:accuracy}), longer than the ones obtained with the selective techniques.

For selective techniques, the observations are consistent with previous reports~\citep{FradyCapacity2018, HerscheHDMClassifier2021} where LR decoding (red solid lines) improves over Codebook decoding (blue solid lines) for small values of $D$ (e.g., $D=5$) but the improvement diminishes as $D$ increases (cf. rightmost panels in Figure~\ref{fig:perf:noiseless}).

As for feedback-based techniques, it is clear that EA (dashed lines) extended the high-fidelity regime for the corresponding selective techniques. 
At the same time, there was a critical value of the accuracy of a selective technique after which the accuracy of EA reduced drastically (since incorrect predictions added noise rather than removing it). 
The use of MP (dotted lines) partially alleviated this issue, as the crosstalk noise was removed symbol after symbol. 
This resulted in even longer high-fidelity regimes and a more gradual transition from the high-fidelity to the low-fidelity regimes (where performance was near chance).

In order to compare the computational complexity of different techniques, we measured the average number of floating point operations (flops) per decoding of a symbol using the PAPI (Performance Application Programming Interface) library~\citep{TerpstraPAPI2010} (Figure~\ref{fig:perf:noiseless} lower panels).
Not surprisingly, selective techniques (especially Codebook decoding) were the cheapest to compute.\footnote{
Out of full fairness, we should note that we have not included the one-time cost of computing $\tilde{\mathbf{C}}$ for techniques that used LR decoding, which would surely add to the already reported costs. 
}
The key observation to make, however, is that the techniques that provided the highest information rate (e.g., the hybrid techniques) also required the largest number of computations.\footnote{
It should be noted further that the values reported in Figure~\ref{fig:perf:noiseless} assume that the whole sequence needs to be decoded.
In the case when only a single symbol of the sequence should be decoded, the gap between selective and complete decoding techniques will be even larger since the techniques for complete decoding would need to decode the whole sequence anyway while the techniques for selective decoding would be able to decode an individual field without accessing the other ones.
}
This observation suggests that there is a tradeoff between the computational complexity of a decoding technique and the amount of information it can decode from a distributed representation.
As an example, we can consider techniques using EA (dashed lines) and MP (dotted lines). 
It was already noted above that the use of MP noticeably improves the high-fidelity regime.   
However, there is a computational price to be paid for the improvement since EA involves up to $vr$ repetitions (grows linearly with $v$) of some selective decoding while MP using EA as a part of its algorithm requires up to $v(v+1)r/2$ repetitions (grows quadratically with $v$), which contributes substantially to the computational complexity (cf. the corresponding curves in  the lower panels in Figure~\ref{fig:perf:noiseless}). 

\subsection{Noisy decoding}
\label{sect:empirical:noisy}

\begin{figure}[h!]
    \centering
    \includegraphics[width=1.0\columnwidth]{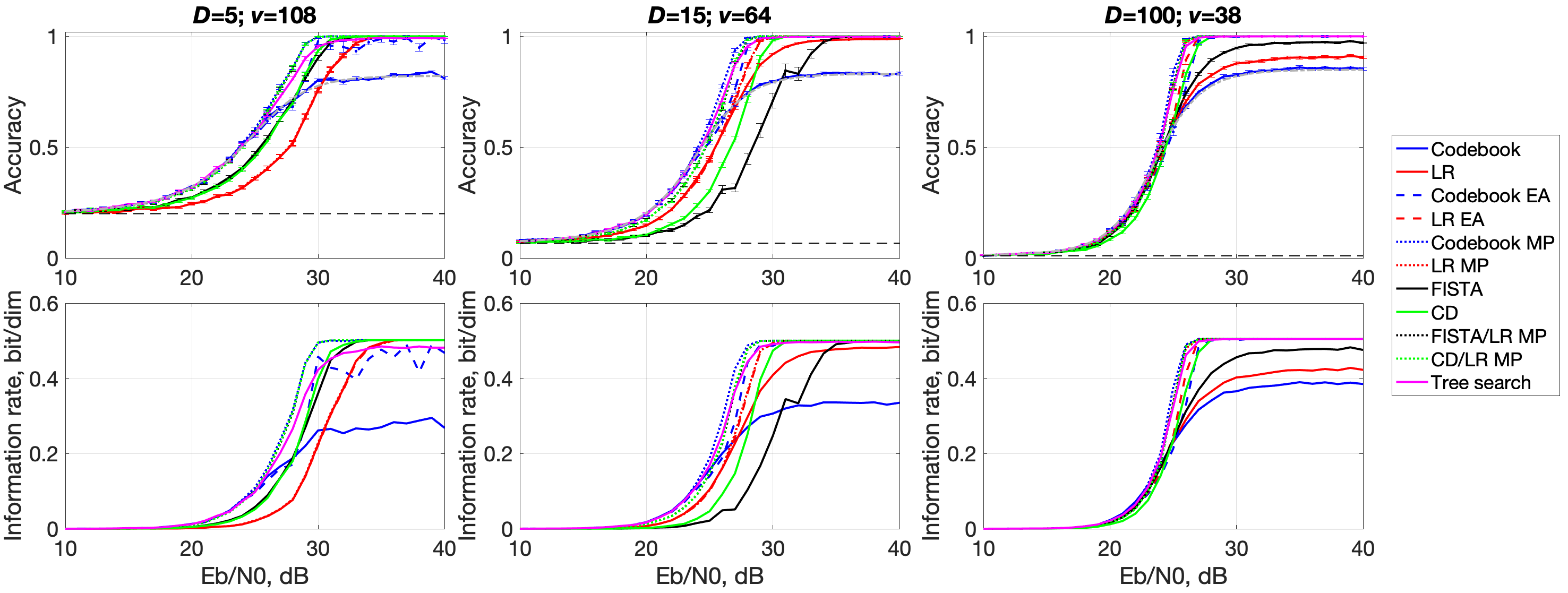}
    \caption{
    The decoding accuracy and information rate against the amount of external noise added to $\mathbf{y}$.
    The upper panels depict accuracy while the lower panels correspond to information rate. 
    For each value of $D$, the value of $v$ was chosen such that the coding rate $v\log_2(D) /n$ would be approximately 0.5.
    The reported results are averages obtained from twenty randomly initialized codebooks. 
    Ten random sequences were simulated for each codebook per each value of Eb/N0.   
    In the upper panels, bars depict $95\%$ confidence intervals, thin dashed black lines indicate the corresponding random guess at $1/D$; while gray dashed lines correspond to analytical predictions for Codebook decoding.  
    }
    \label{fig:perf:noise}
\end{figure}

In the previous experiment, no external noise was added to $\mathbf{y}$. 
However, in many scenarios distributed representations are exposed to noisy environments, e.g., during data transmission. Therefore, it is worth investigating the behavior of the considered decoding techniques in the presence of noise.
We performed the experiments with the same setup as in the previous experiment using additive white Gaussian noise (AWGN).
In order to account for varying magnitude of $\mathbf{y}$, the normalized signal-to-noise ratio (known as Eb/N0) was used to regulate the amount of noise being added to $\mathbf{y}$ as:
\noindent
\begin{equation}
    \hat{\mathbf{y}}=\mathbf{y} + \mathbf{\varepsilon},
    \label{eq:noise}
\end{equation}
\noindent
where $\mathbf{\varepsilon}$ is an $n$-dimensional vector with AWGN that is randomly sampled for the given signal-to-noise ratio. 
The signal-to-noise ratio for a chosen Eb/N0 was computed as $ \text{Eb/N0} + 10\log_{10}(v \log_2(D)/n)$ and the power of the signal was $\langle \mathbf{y}^2 \rangle$.
The noisy version of $\mathbf{y}$, $\hat{\mathbf{y}}$, was an input for all the decoding techniques. 
Figure~\ref{fig:perf:noise} reports the results. 
Note that for this experiment we have considered an additional decoding technique, labeled ``Tree search'' (with magenta solid lines). 
It was introduced in~\citep{HsuNonOrthogonalModulation2020} where the problem of decoding individual symbols from the superposition hypervector was formulated as an optimization problem (similar to the LASSO techniques) but instead using a tree-based search to iteratively decode each symbol. 
Instead of concatenating one-hot encodings of symbols into a single vector to do the joint optimization according to Eq.~\ref{eq:reg:lasso}, the technique in~\citep{HsuNonOrthogonalModulation2020} searches individual symbols one by one and keeps track of  $K$ best candidates for each symbol.
We implemented the tree-based search with the universal sorting keeping $K=2$ best candidates at each step.

In Figure~\ref{fig:perf:noise}, for each value of $D$, the value of $v$ was chosen to match the information rate of $0.5$ bits per dimension.
Clearly, if the amount of added noise was too high, the accuracies were down to random guess values ($1/D$) so no information was retrieved, hence, the information rate was zero. 
Once signal-to-noise improved, each decoding technique reached its highest accuracy matching the corresponding noiseless value (cf. Figure~\ref{fig:perf:noiseless}). 
The tree-based search included in the comparison, as expected, performed better than the EA-based techniques but slightly worse than 
the hybrid techniques.
Also, with the increased value of $D$ (cf. rightmost panels), the difference in the performance of different techniques during the transition from the low-fidelity regime to the high-fidelity regime was not significant. 
This was, however, not the case for lower values of $D$ where the first thing to notice was that Codebook decoding was the first technique to demonstrate accuracies that were  higher than the random guesses $1/D$. 
As a consequence, feedback-based techniques using Codebook decoding also performed well, for example,  the Codebook MP decoding was either on a par or better than the CD/LR~MP decoding. 
Thus, it is useful to keep in mind that in some scenarios rather simple decoding techniques might be still worthwhile  in terms of both performance and computational cost.

\subsection{Decoding with limited precision}
\label{sect:results:clipping}

Before, we assumed that the superposition operation used when forming $\mathbf{y}$ was linear. 
A practical disadvantage of this assumption is that for large values of $v$, a possible range of values of $\mathbf{y}$ is also large making it expensive to store $\mathbf{y}$. 
Therefore, it is practical to consider limiting the precision of components of $\mathbf{y}$. 
Since $\mathbf{y}$ consists of only integer values, a clipping function that is commonly used in neural networks~\citep{FradyCapacity2018, KleykointSOM2019,KleykoDensityEncoding2020} is a simple choice to keep the values of components of $\mathbf{y}$ in a limited range that is regulated by  a threshold value (denoted as $\kappa$):
\noindent
\begin{equation}
f_\kappa (y) = 
\begin{cases}
-\kappa & y \leq -\kappa \\
y & -\kappa < y < \kappa \\
\kappa & y \geq \kappa
\end{cases}
\label{eq:clipping}
\end{equation}
\noindent
Thus, when using the clipping threshold $\kappa$, the values of $f_\kappa (\mathbf{y})$ will be integers in the range between $-\kappa$ and $\kappa$.
In this case, each neuron can be represented using only $\log_2(2\kappa+1)$ bits of memory.
For example, when $\kappa=3$, there are seven unique values that can be stored with just three bits.

\begin{figure}[t]
    \centering
    \includegraphics[width=1.0\columnwidth]{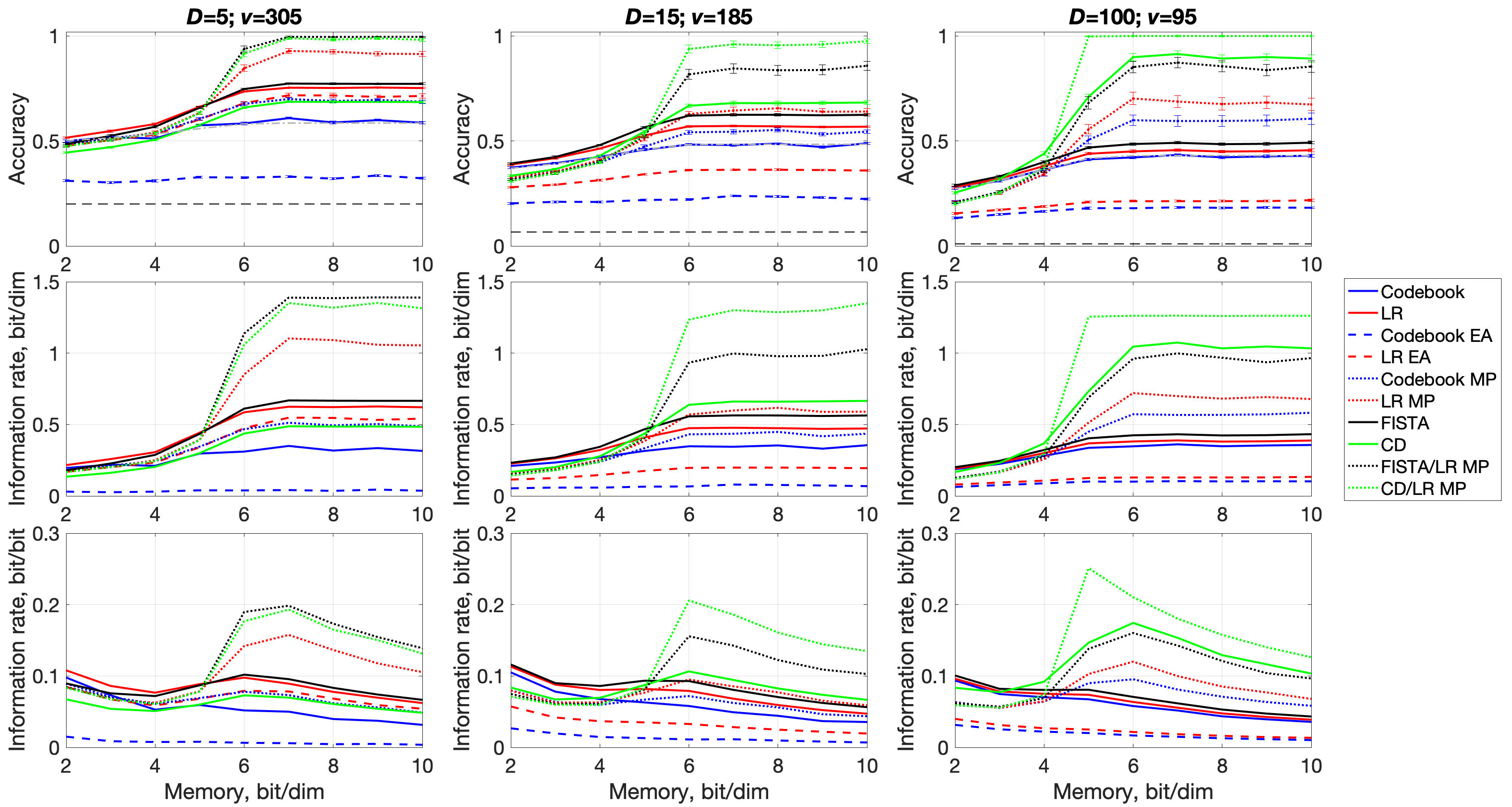}
    \caption{
    The decoding accuracy and information rate against the limited precision used to store $\mathbf{y}$.
    The upper panels depict accuracy, the middle panels correspond to the information rate in bits per dimension, and the lower panels show the information rate in information bits per storage bits. 
    The reported results are averages obtained from twenty randomly initialized codebooks. 
    Ten random sequences were simulated for each codebook per each memory size.   
    In the upper panels, bars depict $95\%$ confidence intervals, thin dashed black lines indicate the corresponding random guess at $1/D$; while gray dashed lines correspond to analytical predictions for Codebook decoding.     
    }
    \label{fig:perf:clipping}
\end{figure}

In order to investigate the effect of the limited precision, we used the clipping function with the following values of $\kappa \in \{1,  3,  7, 15, 31, 63, 127, 255, 511 \}$ that approximately required $[2:10]$ bits of storage per dimension, respectively. 
The results are reported in Figure~\ref{fig:perf:clipping}, where the upper panels show the accuracy, while the middle and the lower panels depict the information rate in bits per dimension and bits per storage bits, respectively. 
The values of $v$ for each $D$ were chosen to match their peaks of the information rate as observed in Figure~\ref{fig:perf:noiseless}.
In order to account for the effect of the reduced scaling due to the use of the clipping function, the clipped representations were re-scaled by a constant factor, which was estimated analytically based on $v$ and $\kappa$, to match the power of the original representations. 

First, it is clear that making the superposition operation to be nonlinear was detrimental for some of the decoding techniques.  
This is particularly true for EA techniques that performed worse than their corresponding selective techniques. 
EA was so sensitive to the use of the clipping function as it is based on the assumption that the superposition operation to form $\mathbf{y}$ is linear. 
The other techniques, however, managed to get close to their information rate in bits per dimension (middle panels in Figure~\ref{fig:perf:clipping}; cf. middle panels in Figure~\ref{fig:perf:noiseless}) once $\kappa$  was sufficiently large.
We could also see that the best information rate in information bits per storage bits (lower panels) was achieved for the smallest value of $\kappa$ that allowed reaching the maximum of the information rate in bits per dimension. 
Note also that for larger values of $D$, the information rate in information bits per storage bits was higher. 
This is the case as for lower values of $D$ the peak in the information rate was observed for higher values of $v$ that in turns means large range of values of $\mathbf{y}$ and, hence, larger values of $\kappa$ to preserve most of the range. 
Finally, it is worth noting that the second smaller peak in the information rate in information bits per storage bits was observed for the smallest value of $\kappa=1$ where the selective techniques were the best option. 
This is expected since the range is so limited that neither feedback-based nor LASSO techniques can benefit from it.

\section{Discussion}
\label{sect:discussion}

\subsection{Summary of the study}

In this paper, we focused on the problem of retrieving information from compositional distributed representations obtained using the principles of HD/VSA.
To the best of our knowledge, this is the first attempt to survey, categorize, and quantitatively compare decoding techniques for this problem.
Our taxonomy reveals that decoding techniques from other research areas can be utilized, such as reservoir computing, sparse signal representation, and communications. 
In fact, some of the investigated techniques were not used previously in HD/VSA, but improved the information rate bounds beyond the state-of-the-art.
We also introduced a novel decoding technique -- matching pursuit with explaining away (Section~\ref{sect:decoding:complete:feed:mp}).
It should be noted that the experiments in this study used the Multiply-Add-Permute model.
While it was shown before~\cite{FradyCapacity2018, SchlegelVSAComparison2020} that for Codebook decoding some HD/VSA models are more accurate (e.g., Fourier Holographic Reduced Representations model~\cite{PlateHolographic1995}) this observation should not affect the consistency of relative standing of the considered decoding techniques when evaluated on models other than Multiply-Add-Permute.
Our decoding experiments explored three different encoding scenarios: the hypervector formed by plain linear superposition, linear superposition with external noise, and lossy compression of linear superposition using component-wise clipping.
The standard decoding technique in HD/VSA -- Codebook decoding -- was in all scenarios outperformed by other techniques. 
But, nevertheless, it combines decent decoding performance with other advantages: absence of free parameters that require tuning and the lowest computational complexity. 
In the first scenario of linear superposition with no external noise, LASSO techniques performed exceptionally well. 
In other scenarios, high noise, or compression with strong nonlinearity (cf. Figure~\ref{fig:perf:clipping}), the assumptions in the optimization approach (\ref{eq:reg:lasso}) are violated and, accordingly, the performance is worse than with simpler techniques.
Notably, in our experiments, the hybrid decoding techniques combining LASSO techniques with matching pursuit with explaining away advanced the theory of HD/VSA by improving the information rate bounds of the distributed representations reported before in~\citep{FradyCapacity2018,HerscheHDMClassifier2021} by at least $17$\%.
However, this improvement comes at the price of performing several orders of magnitude more operations compared to the simplest selective techniques (cf. lower panels in Figure~\ref{fig:perf:noiseless}, which highlights the tradeoff between the computational complexity and information rate).

\subsection{Related work}

\subsubsection{Randomized neural networks and reservoir computing}

Decoding from distributed representations can be seen as a special case of function approximation, which connects it to randomized neural networks and reservoir computing~\citep{RCNNSsurvey}.
As we highlighted in Section~\ref{sect:decoding:readout:cov}, this interpretation allows learning a readout matrix for each position in a sequence from training data.
This technique was introduced to HD/VSA in~\citep{FradyCapacity2018} that has also shown that when distributed representations are formed according to (\ref{eq:perm_seq}), the readout matrices do not have to be trained -- they can be computed using the covariance matrix from (\ref{eq:cov:fine}).

\subsubsection{Sparse Coding (SC) and Compressed Sensing (CS)}

As indicated in Section~\ref{sect:decoding:optimization}, the task of retrieving from (\ref{eq:perm_seq}) can be framed as sparse inference procedure used within SC~\citep{OlshausenEmergence1996} and CS~\citep{DonohoCompressed2006}. 
Within HD/VSA literature this connection was first made in~\citep{Summers2018} for decoding from sets and in~\citep{FradySDR2020} for sets and sequences. Similar to SC and CS, L0 sparsity is more desirable than L1 sparsity since the sparse vector, $\mathbf{x} \in \{0,1\}^{vD}$, is composed of variables which are exactly zero and one. In general, optimization of the L0 penalty is a hard problem. Optimization with the L1 penalty thresholds small values leading to a sparse vector with many variables being zero. Efficient algorithms exist for optimization of the L1 penalty, which provides a practical technique for performing sparse inference.

\subsubsection{Communications}

The problem of decoding individual messages from their superposition as in (\ref{eq:perm_seq}) is the classical multiple access channel (MAC) problem in communications. The capacity region which specifies the achievable rates for all users given their signal-to-noise ratios has been fully characterized ~\citep{cover1999elements}. It is known that the capacity region of a MAC can be achieved by code-division-multiple access (CDMA), where separate codes are used by different senders and the receiver decodes them one by one. This so-called \textit{successive interference cancellation} (onion-peeling) is the key idea used for Codebook decoding with EA. Understanding how close the performance of the decoder is to the capacity will provide us insights in improving decoder design in the future.

Within HD/VSA, the decoding with inference cancellation (i.e., EA) was introduced in~\citep{KimHDM2018} that proposed to combine forward error correction and modulation using Fourier Holographic Reduced Representations model~\citep{PlateHolographic1995}.
Similar to the results reported here, the main motivation for using inference cancellation was the fact that it significantly improved the quality of decoding compared to Codebook decoding.  
Later in~\citep{HerscheHDMClassifier2021}, the so-called ``soft-feedback'' technique was introduced, similar to MP it makes use of the prediction's confidence.
Another improvement on top of EA was the tree-based search~\citep{HsuNonOrthogonalModulation2020} mentioned above. It outperformed EA-based techniques with the caveat that the complexity of the tree-based search grows exponentially with the number of branches, so only the $K$ best candidates for each symbol were retained~\citep{HsuNonOrthogonalModulation2020}. 
This imposes a trade-off between decoding accuracy and computation/time complexity. 
It is also prone to errors when several candidates share the same score.

Another development within communications that is very similar to the retrieval task considered here is Sparse Superposition Codes (SSC)~\citep{barron2010toward}. 
SSCs are capacity achieving codes with a sparse block structure that are closely related to SC~\citep{OlshausenEmergence1996} and CS~\citep{DonohoCompressed2006}. 
SSC decoding algorithms are like those studied in this work, e.g., L1 minimization, successive interference cancellation, or approximate belief propagation techniques. Future work should investigate SSC constructed from the encoding and decoding strategies from this work.

\subsubsection{Related work within HD/VSA literature}

\label{sec:related:VSA}

Besides the works mentioned above that have been connecting the task of retrieving from distributed representations formed by HD/VSA to tasks within other areas, there are also works that have studied the Codebook decoding technique. 
Early analytical results on the performance of Codebook decoding with real-valued hypervectors  were given in~\citep{PlateHolographic2003}.
For the case of dense binary hypervectors, an important step for obtaining the analytical results is in estimating an expected Hamming distance between  the compositional hypervector and a symbol's hypervector (see, e.g., expressions in~\citep{KanervaFully1997, MitrokhinSensorimotor2019, KleykoCommentariesSR2020}).
Further steps for the dense binary/bipolar hypervectors were presented in~\citep{GallantRepresenting2013, KleykoHolographic2017, RahimiNanoscalable2017}. 
The performance in the case of sparse binary hypervectors~\citep{RachkovskijStructures2001} was analyzed in~\citep{KleykoTradeoffs2018}.
The most general and comprehensive analytical studies of the performance of Codebook decoding for different HD/VSA models were recently presented in~\citep{FradyCapacity2018,KleykoPerceptron2020} 
while other recent studies~\citep{ThomasHDFoundations2021,ClarksonVSACapacity2023} have provided theoretical bounds of several HD/VSA models in other scenarios. 
Some recent empirical studies of the capacity of HD/VSA can be also found in~\citep{MirusCapacity2020, SchlegelVSAComparison2020}.

Finally, it is worth noting that the problem formulation considered here is very similar to the trajectory association task that was proposed in~\citep{PlateRecurrent1992} and can be used to study the memory capacity of recurrent neural networks (see, e.g., ~\citep{FradyCapacity2018,DanihelkaAssociative2016}).

\subsection{Future work}

In this study, we have surveyed the key ideas and techniques to solve the retrieval task. There are, however, more specific techniques to try, as well as other angles one can choose to look at the problem. 
A possibility for future work is to compare the computation complexity and decoding accuracy between different LASSO techniques.
Some other techniques that we have not simulated but are worth exploring include  
MP with several iterations, genetic algorithms for refining the best current solution, and LASSO solving for a range of values of $\lambda$~(see~\citep{Summers2018} for some experiments within HD/VSA).

While in this study, we have fixed the formation of distributed representations (Eq.~(\ref{eq:perm_seq}) in Appendix~\ref{sec:problem}), it is expected that the choice of the transformation of input data can affect the performance of the decoding techniques. 
For instance, as we saw in Figure~\ref{fig:perf:noiseless} working with smaller codebooks leads to increased information rate. 
Another notable example of importance of input transformation are fountain codes~\citep{mackay2005fountain}, where the packet distribution can be optimized to minimize the probability of error.
Therefore, as a part of future work, we also plan to consider other transformations to distributed representations not considered here.


\begin{appendices}

\section{\large Background}
\label{sect:background}

\subsection{Vector Symbolic Architectures}
\label{sect:supp:vsa}

In this section, we provide a summary from~\citep{KleykoComputingParadigm2021} to briefly introduce HD/VSA~\citep{KanervaHyperdimensional2009, GaylerJackendoff2003}\footnote{
Please consult~\citep{KleykoSurveyVSA2021Part1,KleykoSurveyVSA2021Part2} for a comprehensive survey of HD/VSA. 
}
using the Multiply-Add-Permute (MAP) model~\citep{GaylerJackendoff2003}  to showcase a particular HD/VSA realization. 
It is important to keep in mind that HD/VSA can be formulated with different types of vectors, namely those containing real~\citep{PlateHolographic1995, GallantRepresenting2013}, complex~\citep{PlateHolographic1995}, or binary entries~\citep{KanervaFully1997,LaihoSparse2015, RachkovskijStructures2001, FradySDR2020}.
The key components of any HD/VSA model are:
\noindent
\begin{itemize}
    \item High-dimensional space (e.g., integer; $n$ denotes the dimensionality);
    \item Pseudo-orthogonality (between two random vectors in this high-dimensional space);
    \item Similarity measure (e.g., dot (inner) product or cosine similarity);
    \item Atomic representations (e.g., random i.i.d.  high-dimensional vectors, a.k.a. hypervectors);
    \item Item memory storing atomic hypervectors and performing auto-associative search;
    \item Operations on hypervectors.
\end{itemize}
\noindent

In MAP~\citep{GaylerJackendoff2003}, the atomic hypervectors are bipolar random vectors, where each vector component is selected randomly and independently from $\{-1,+1\}$.
These random i.i.d. atomic hypervectors can serve to represent ``symbols'' in HD/VSA (i.e., categorical objects), since such vectors are pseudo-orthogonal to each other (due to the concentration of measure phenomenon) and, thus, are treated as dissimilar. 

Additionally, each HD/VSA model defines three key operations used to manipulate atomic hypervectors, we specify their implementations in MAP: 
\noindent
\begin{itemize}
    \item Superposition a.k.a. bundling (denoted as $+$; implemented as component-wise addition possibly followed by some normalization function); 
    \item Permutation (denoted as $\rho$; implemented as a rotation of components);
    \item Binding (denoted as $\odot$; implemented as component-wise multiplication a.k.a. Hadamard product; not used in this paper).    
\end{itemize}
\noindent

HD/VSA models also need to define a similarity measure between two vector representations. 
In this paper, for this purpose we will use dot product and cosine similarity that are computed for two hypervectors $\mathbf{a}$ and $\mathbf{b}$ as: 
\noindent
\begin{equation}
\label{eq:inner} 
\mathbf{a}^\top \mathbf{b}
\end{equation}
\noindent
and
\noindent
\begin{equation}
\label{eq:cosine} 
\frac{\mathbf{a}^\top \mathbf{b}}{||\mathbf{a}||_2||\mathbf{b}||_2},
\end{equation}
\noindent
respectively; $||\cdot||_2$ denotes L2 norm of a hypervector.

\subsection{Problem formulation}
\label{sec:problem}

In this section, we present a transformation (i.e., an encoding scheme) that is used to form distributed representations for this study. 
It is worth noting that the use of HD/VSA operations allows forming compositional distributed representations for a plethora of data structures such as sets~\citep{KanervaHyperdimensional2009, KleykoABF2020}, sequences~\citep{KanervaHyperdimensional2009, HannaganHolographic2011, ThomasHDFoundations2021},  state machines~\citep{OsipovHD_FSA2017, YerxaUCBHD_FSA2018}, hierarchies, or predicate relations~\citep{RachkovskijStructures2001, PlateHolographic2003,GallantVSAJSON2022}, etc. (please consult~\citep{KleykoComputingParadigm2021} for a detailed tutorial on representations of these data structures). 

For the sake of focusing on decoding techniques, we use only one simple but common transformation for representing a symbolic sequence of length $v$. 
A sequence (denoted as $\mathbf{s}$, e.g., $\mathbf{s}=(a,b,c,d,e)$) is assumed to be generated randomly. 
Symbols constituting the sequence are drawn from an alphabet of finite size $D$ and the presence of each symbol in any position of the sequence is equiprobable.

In order to form a distributed representation of a sequence, first we need to create an item memory, $\mathbf{\Phi}$, (we call it the \textit{codebook}) that stores atomic $n$-dimensional random i.i.d. bipolar dense hypervectors corresponding to symbols of the alphabet\footnote{
For the sake of simplicity let us assume that the symbols are integers between $0$ and $v-1$. This notation makes it more convenient to introduce the decoding techniques in Section~\ref{sect:decoding}. 
}, thus, $\mathbf{\Phi}  \in \{-1,1\}^{n \times D}$.
The hypervector of the $i$th symbol in $\mathbf{s}$ will be denoted as $\mathbf{\Phi}_{\mathbf{s}_i}$.
It should be noted that in HD/VSA it is a convention to draw hypervectors' components randomly unless there are good reasons to make them correlated (see~\citep{FradyFunctions2021,FradyFunctionsNICE2022}).
The presence of correlation will, however, reduce the performance of the decoding techniques because their signal-to-noise ratio will be lower and, thus, the decoding becomes harder.

For sequence transformations, there is a need to associate a symbol's hypervector with a symbol's position in the sequence. 
There are several approaches to do so; we will use one that relies on the permutation operation~\citep{PlateSequences1995, SahlgrenOrder2008, KanervaHyperdimensional2009, KleykoPermuted2016, FradyCapacity2018, KanervaComputing2019}.
The idea is that before combining together the hypervectors of sequence symbols, the position $i$ of each symbol is associated by applying some fixed permutation $v-i$ times to its hypervector\footnote{ 
It is worth pointing out that the reverse order of applying successive powers of a permutation can be used as well.
}
(e.g., $\rho^2(\mathbf{\Phi}_c)$ for the sequence above). 

The last step is to combine the sequence symbols into a compositional hypervector (denoted as $\mathbf{y}$) representing the whole sequence. 
We do it using the superposition operation. 
For the sequence above the compositional hypervector is: 
\noindent
\begin{equation*}
\mathbf{y} = \rho^4(\mathbf{\Phi}_a) + \rho^3(\mathbf{\Phi}_b) + \rho^2(\mathbf{\Phi}_c) + \rho^1(\mathbf{\Phi}_d) + \rho^0(\mathbf{\Phi}_e). 
\end{equation*}
In general, a given sequence $\mathbf{s}$ of length $v$ is represented as: 
\noindent
\begin{equation}
\label{eq:perm_seq}
\mathbf{y} = \phi(\mathbf{s}) = \sum_{i=1}^{v}\rho^{v-i}( \mathbf{\Phi}_{\mathbf{s}_i}).
\end{equation}
\noindent
Note that a transformation of the sequence similar to the one in Eq.~(\ref{eq:perm_seq}) can be obtained by assigning $v$ random bipolar hypervectors (one for each unique position) and using the binding operation on the corresponding hypervectors to represent the association between a symbol and its position. 
We, however, do not report the experiments on such a representation since in terms of decoding performance it is equivalent to that in Eq.~(\ref{eq:perm_seq}).  
In this study, we also assume that the superposition operation is linear unless noted otherwise (see 
Section~\ref{sect:results:clipping}), i.e., that no normalization operation is applied to the result of~(\ref{eq:perm_seq}).

The problem of decoding from compositional distributed representation $\mathbf{y}$ is formulated as follows: for given $v$, $\mathbf{\Phi}$, and $\mathbf{y}$ the task is to provide a reconstructed  sequence (denoted as $\hat{\mathbf{s}}$) such that $\hat{\mathbf{s}}$ is as close as possible to the original sequence $\mathbf{s}$.

\subsection{Performance metrics}
\label{sec:back:perf}

In this section, we introduce two performance evaluation metrics used in this study to assess the quality of the reconstructed sequence $\hat{\mathbf{s}}$. 

\subsubsection{Accuracy of the decoding}
\label{sec:back:perf:accuracy}

For the decoding of a sequence's symbols from its compositional distributed representation $\mathbf{y}$, the accuracy of the correct decoding (denoted as $a$) is a natural metric to characterize the performance of a decoding technique.  
Since the superposition operation used in this study is linear,
the accuracy of decoding symbols in different positions of the sequence is the same, and so there is no need to compute separate accuracies for different positions. 
The accuracy is computed empirically over $g$ randomly generated sequences, using the predictions in the reconstructed sequence $\hat{\mathbf{s}}$ (obtained by some decoding technique) and the original sequence $\mathbf{s}$: as an average ratio of symbols, which were decoded correctly:
\noindent
\begin{equation}
\label{eq:accuracy}
\begin{split}
a &= \frac{1}{gv}\sum_{n=1}^{g} \sum_{i=1}^{v} [\mathbf{s}_i^{(n)}= \hat{\mathbf{s}}_i^{(n)}]
\end{split}
\end{equation}
\noindent
where $[\cdot]$ is the indicator function set to one when the symbols match and to zero otherwise. 
Note that since symbols in sequence $\mathbf{s}$ are equiprobable, the accuracy of a random guess is $1/D$.

\subsubsection{Information decoded from distributed representation}
\label{sec:information}

The accuracy is an intuitive performance metric for the task of sequence decoding and allows characterizing performance of various decoding techniques.
Its disadvantage, however, is that the accuracy also depends on the dimensionality of representation, $n$, size of the codebook, $D$, and number of symbols in the sequence, $v$.
In order to, e.g., study the effect of these parameters on distributed representation, it is convenient to use a single scalar metric characterizing the quality of decoding that would take into account particular values of $n$, $D$, and $v$.
For this purpose, we use the total amount of information decoded per a dimension of representation.  
The amount of information decoded for a single symbol (denoted as $I_{\mathrm{symb}}$) is calculated using the corresponding accuracy and size of the codebook as\footnote{
We do not go into the detailed derivation of this equation here. Please refer to Section 2.2.3 in~\citep{FradyCapacity2018} for the details.
}:  
\noindent
\begin{equation}
\label{eq:MI:item}
\begin{split}
I_{\mathrm{symb}}(a, D) = & a \log_2(D a ) +  (1-a) \log_2 \left( \frac{D}{D-1} (1-a) \right).
\end{split}
\end{equation}
\noindent
Note that when the accuracy equals that of a random guess ($1/D$) the amount of decoded information would be $0$.
The total amount of information decoded from the distributed representation is calculated as a sum of information extracted for all symbols: 
\noindent
\begin{equation}
\label{eq:MI:tot}
I_{\mathrm{tot}}=\sum_{i=1}^{v} I_{\mathrm{symb}}(a, D)= vI_{\mathrm{symb}}(a, D).
\end{equation}
\noindent
Finally, since $I_{\mathrm{tot}}$ does not account for the dimensionality of representation, it makes sense to consider the amount of decoded information per a single dimension:
\noindent
\begin{equation}
\label{eq:MI:dim}
I_{\mathrm{dim}}=\frac{I_{\mathrm{tot}}}{n}.
\end{equation}
\noindent
Thus, $I_{\mathrm{dim}}$ corresponds to an information rate that is a relative measure in bits per dimension that accounts for all three parameters $n$, $D$, and $v$ and can be used for a fair comparison of various transformations using different choices of these parameters.

\section{\large Additional experiments against the dimensionality of representations}
\label{sect:dimensionality}

\begin{figure}[t]
    \centering
    \includegraphics[width=1.0\columnwidth]{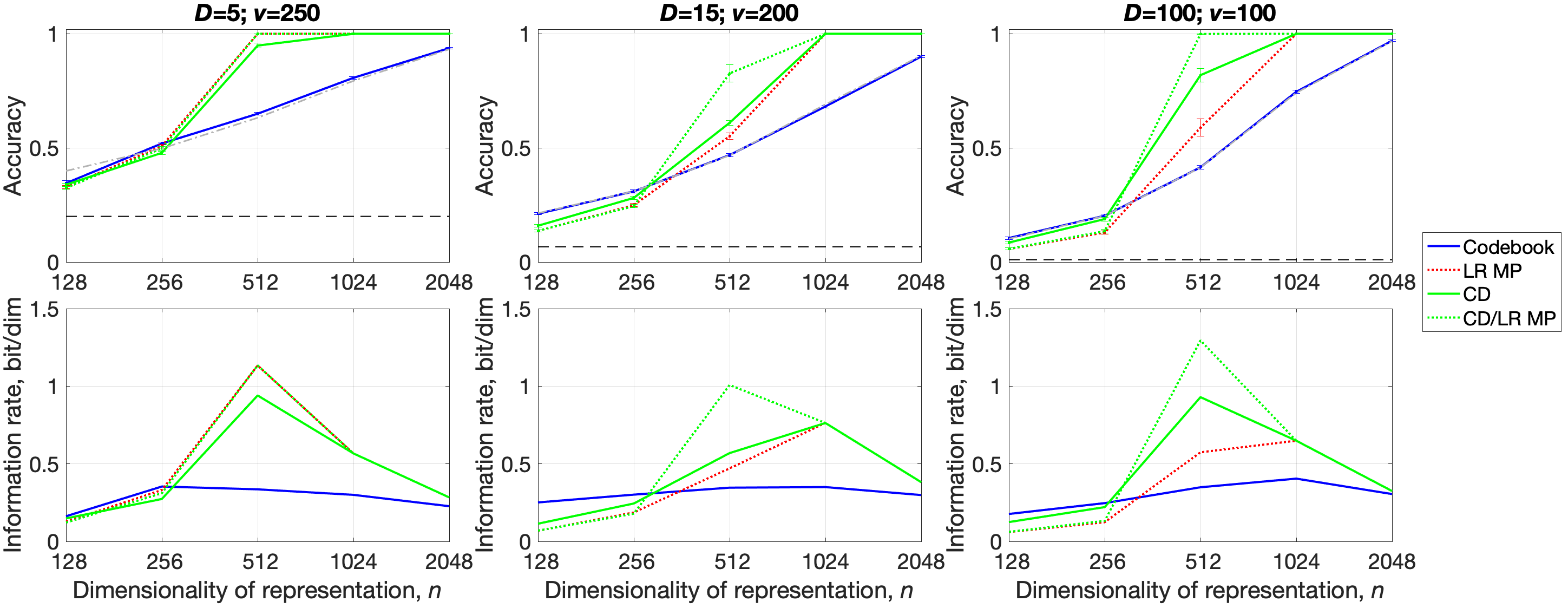}
    \caption{
    The decoding accuracy and information rate against dimensionality of representations $n$ for three different codebook sizes $D$ using a subset of the considered decoding techniques. 
    The upper panels depict accuracy, while the lower panels correspond to information rate.    
    The reported results are averages obtained from ten randomly initialized codebooks. 
    Ten random sequences were simulated for each codebook per each $v$.
    In the upper panels, bars depict $95\%$ confidence intervals, thin dashed black lines indicate the corresponding random guess at $1/D$; while gray dashed lines correspond to analytical predictions for Codebook decoding.      
    }
    \label{fig:perf:dimen}
\end{figure}

In the main text, the dimensionality of representations was fixed to $n=500$.
While it is well-known~\cite{FradyCapacity2018} that with the increased dimensionality the decoding accuracy will also increase, it is still worth demonstrating this effect empirically within the experimental protocol on this study.
To do so, we hand-picked four decoding techniques from the considered groups: ``Codebook'', ``LR MP'', ``CD'', and ``CD/LR MP''. 
The techniques were evaluated using the values of $n$ from $\{128,  256,  512, 1024, 2048 \}$.
For each codebook size, the choice of $v$ was made qualitatively based on Figure~\ref{fig:perf:noiseless} choosing various relative performance gaps (at $n=500$) between the techniques. 
The results are reported in Figure~\ref{fig:perf:dimen}, where the upper panels depict the accuracy, while the lower panels show the information rate. 
There are no surprising observations about Figure~\ref{fig:perf:dimen}. 
Eventually, given a sufficient dimensionality, each technique entered the high-fidelity regime.
The relative ordering of the techniques persisted with increased $n$ (unless all were perfectly accurate), while for lower  $n$ crosstalk noise is too high, effectively similar to adding a lot of noise as in Figure~\ref{fig:perf:noise} where all techniques performed equally poorly.    

\section{\large Pseudo-code for hybrid techniques}
\label{sect:pseudocode}

Section~\ref{sect:decoding:hybrid} stated that it is worth combining primitives from several various techniques, which results in hybrid techniques. 
In particular, the results in Section~\ref{sect:empirical} featured two such techniques: the combination of CD decoding with LR decoding with MP (``CD/LR MP'') as well as the combination of FISTA decoding with LR decoding with MP (``FISTA/LR MP''). 
In order to provide more details on the techniques, Algorithm~\ref{alg:hybrid} lists the corresponding pseudo-code for ``CD/LR MP''. 
The algorithm for ``FISTA/LR MP'' is not shown explicitly as it is obtained simply by replacing every usage of CD decoding in Algorithm~\ref{alg:hybrid} by FISTA decoding.
The notations used in the algorithm correspond to the ones introduced above. There are, however, several novel notations such as $\text{sim}_{\text{cos}}(\cdot,\cdot,)$ to denote cosine similarity (see Section~\ref{sect:supp:vsa}), $\mathrm{\mathbf{confidence}}$ for a $v$-dimensional vector storing a confidence score for each position in the sequence (see Section~\ref{sect:decoding:complete:feed:mp}),  
$\mathrm{\mathbf{fixed}}$ for a set storing positions with the fixed prediction (see Section~\ref{sect:decoding:complete:feed:mp}), as well as $\texttt{CD}()$ and $\texttt{LRMP}()$ denoting routines for performing CD and LR MP decoding, respectively. 
Note that these routines can take $\mathrm{\mathbf{fixed}}$ as input, which means that the predictions in the corresponding positions will remain unchanged and values of confidence scores in these positions will be set to $-1$ to avoid choosing them again during the $\argmax()$ step.

\newpage
\begin{algorithm}
\caption{A high-level pseudo-code for ``CD/LR MP'' decoding.}
\label{alg:hybrid}
\begin{algorithmic}
\Require $\mathbf{y}$;  $\mathbf{\Phi}$;  $v$; 

\State $\hat{\mathbf{s}}^{\mathrm{CD}}, \mathrm{\mathbf{confidence}}^{\mathrm{CD}} \gets  \texttt{CD}(\mathbf{y}, \mathbf{\Phi}, v, \emptyset, \emptyset)$
\State $\hat{\mathbf{s}}^{\mathrm{LRMP}}, \mathrm{\mathbf{confidence}}^{\mathrm{LRMP}} \gets  \texttt{LRMP}(\mathbf{y}, \mathbf{\Phi}, v, \hat{\mathbf{s}}^{\mathrm{CD}}, \emptyset)$

\If{$\text{sim}_{\text{cos}}(\mathbf{y},\phi(\hat{\mathbf{s}}^{\mathrm{CD}})) \geq \text{sim}_{\text{cos}}(\mathbf{y},\phi(\hat{\mathbf{s}}^{\mathrm{LRMP}}))$}
    \State $\hat{\mathbf{s}} \gets \hat{\mathbf{s}}^{\mathrm{CD}} $
    \State $\mathrm{\mathbf{confidence}}\gets \mathrm{\mathbf{confidence}}^{\mathrm{CD}}$
\Else
    \State $\hat{\mathbf{s}} \gets \hat{\mathbf{s}}^{\mathrm{LRMP}} $
    \State $\mathrm{\mathbf{confidence}}\gets \mathrm{\mathbf{confidence}}^{\mathrm{LRMP}}$
\EndIf

\State $\tilde{\mathbf{y}} \gets \mathbf{y} $
\State $\mathrm{\mathbf{fixed}} \gets \emptyset$

\For{$i = 1$ to $v-1$}
    \State $c \gets \argmax(\mathrm{\mathbf{confidence}})$
    \State append $c$ to $\mathrm{\mathbf{fixed}}$ \Comment{Confidence of positions already included in $\mathrm{\mathbf{fixed}}$ is assumed to be equal $-1$} 
    \State $\tilde{\mathbf{y}} \gets \tilde{\mathbf{y}} -  \rho^{v-c}( \mathbf{\Phi}_{\hat{\mathbf{s}}_c})$
    \State $\hat{\mathbf{s}}^{\mathrm{CD}}, \mathrm{\mathbf{confidence}}^{\mathrm{CD}} \gets  \texttt{CD}(\tilde{\mathbf{y}}, \mathbf{\Phi}, v-i,\hat{\mathbf{s}}, \mathrm{\mathbf{fixed}})$
    \State $\hat{\mathbf{s}}^{\mathrm{LRMP}}, \mathrm{\mathbf{confidence}}^{\mathrm{LRMP}} \gets  \texttt{LRMP}(\tilde{\mathbf{y}}, \mathbf{\Phi}, v-i, \hat{\mathbf{s}}^{\mathrm{CD}}, \mathrm{\mathbf{fixed}})$

    \If{$\text{sim}_{\text{cos}}(\mathbf{y},\phi(\hat{\mathbf{s}}^{\mathrm{CD}})) \geq \text{sim}_{\text{cos}}(\mathbf{y},\phi(\hat{\mathbf{s}}^{\mathrm{LRMP}}))$}
        \State $\hat{\mathbf{s}} \gets \hat{\mathbf{s}}^{\mathrm{CD}} $
        \State $\mathrm{\mathbf{confidence}}\gets \mathrm{\mathbf{confidence}}^{\mathrm{CD}}$
    \Else
        \State $\hat{\mathbf{s}} \gets \hat{\mathbf{s}}^{\mathrm{LRMP}} $
        \State $\mathrm{\mathbf{confidence}}\gets \mathrm{\mathbf{confidence}}^{\mathrm{LRMP}}$
    \EndIf
    
\EndFor

\State \Return $\hat{\mathbf{s}}$ 

\end{algorithmic}
\end{algorithm}

\end{appendices}

\vspace{-0.5cm}
\section*{Acknowledgements}

We would like to thank Spencer Kent for sharing with us the implementation of FISTA algorithm. The work of FTS, BAO, CB, and DK  was supported in part by Intel's THWAI program.
The work of BAO and DK was also supported in part by AFOSR FA9550-19-1-0241.
DK has received funding from the European Union's Horizon 2020 research and innovation programme under the Marie Sk\l{}odowska-Curie grant agreement No 839179. 
The work of CJK was supported by the Department of Defense through the National Defense Science \& Engineering Graduate (NDSEG) Fellowship Program.
FTS was supported by Intel and NIH R01-EB026955.

\newpage

\begingroup
\small
\setstretch{1.0}

\bibliographystyle{apalike}
\bibliography{references}

\endgroup

\end{document}